\documentclass{article}
\usepackage{algorithmic}
\usepackage{algorithm}
\usepackage{microtype}
\usepackage{graphicx}
\usepackage{subcaption}
\usepackage{booktabs} 
\usepackage{graphicx}
\usepackage{multirow} 
\usepackage{multicol}
\usepackage{subcaption}
\usepackage{longtable}
\usepackage{booktabs}
\usepackage{makecell}
\PassOptionsToPackage{svgnames}{xcolor}
\usepackage{colortbl}
\definecolor{mygray}{gray}{.9}
\usepackage{hyperref}

\usepackage[accepted]{icml2026}

\usepackage{amsmath}
\usepackage{amssymb}
\usepackage{mathtools}
\usepackage{amsthm}

\usepackage[capitalize,noabbrev]{cleveref}

\theoremstyle{plain}
\newtheorem{theorem}{Theorem}[section]
\newtheorem{proposition}[theorem]{Proposition}

\theoremstyle{definition}

\theoremstyle{remark}

\newcommand\sbr[1]{\left( #1 \right)}
\newcommand\mbr[1]{\left[ #1 \right]}

\icmltitlerunning{Data Agent: Learning to Select Data via End-to-End Dynamic Optimization}

\begin{document}

\twocolumn[
  \icmltitle{Data Agent: Learning to Select \\Data via End-to-End Dynamic Optimization}



  \icmlsetsymbol{equal}{*}

  \begin{icmlauthorlist}
    \icmlauthor{Suorong Yang}{nus,nju,cnrs}
    \icmlauthor{Fangjian Su}{nju}
    \icmlauthor{Hai Gan}{nju}
    \icmlauthor{Ziqi Ye}{nju}
    \icmlauthor{Jie Li}{nju}
    \icmlauthor{Baile Xu}{nju}
    \icmlauthor{Furao Shen}{nju}
    \icmlauthor{Soujanya Poria}{ntu}

  \end{icmlauthorlist}

  \icmlaffiliation{nus}{National University of Singapore}
  \icmlaffiliation{nju}{Nanjing University}
  
  \icmlaffiliation{ntu}{Nanyang Technological University}
  \icmlaffiliation{cnrs}{CNRS@CREATE}

  \icmlcorrespondingauthor{Baile Xu}{xubaile@nju.edu.cn}
  \icmlcorrespondingauthor{Soujanya Poria}{soujanya.poria@ntu.edu.sg}

  \icmlkeywords{Machine Learning, ICML}

  \vskip 0.3in
]



\printAffiliationsAndNotice{}  

\begin{abstract}
Dynamic Data selection aims to accelerate training by prioritizing informative samples during online training.
However, existing methods typically rely on task-specific handcrafted metrics or static/snapshot-based criteria to estimate sample importance, limiting scalability across learning paradigms and making it difficult to capture the evolving utility of data throughout training.
To address this challenge, we propose Data Agent, an end-to-end dynamic data selection framework that formulates data selection as a training-aware sequential decision-making problem.
The agent learns a sample-wise selection policy that co-evolves with model optimization, guided by a composite reward that integrates loss-based difficulty and confidence-based uncertainty signals.
The reward signals capture complementary objectives of optimization impact and information gain, together with a tuning-free adaptive weighting mechanism that balances these signals over training.
Extensive experiments across a wide range of datasets and architectures demonstrate that Data Agent consistently accelerates training while preserving or improving performance, e.g., reducing costs by over 50\% on ImageNet-1k and MMLU with lossless performance.
Moreover, its dataset-agnostic formulation and modular reward make it plug-and-play across tasks and scenarios, e.g., robustness to noisy datasets, highlighting its potential in real-world scenarios. Code is available at \url{https://github.com/Jackbrocp/Data-Agent}.


  
\end{abstract}

\section{Introduction}
Deep learning has made significant progress in recent years, with model architectures becoming increasingly deep and complex to achieve state-of-the-art performance~\cite{llama,gpt-4}.
However, this progress has created a growing demand for ever-larger training datasets, which leads to substantial training costs.
Such costs degrade training efficiency and are usually unaffordable for researchers with limited computational resources.
More importantly, large-scale datasets often contain redundancies, further increasing the training burden without necessarily improving model performance.
To address these and improve data efficiency, data selection methods~\cite{yang2024clip,moderate,dataset_pruning,tdds} aim to identify highly representative subsets of training data, accelerating training without sacrificing performance.
These methods can be broadly categorized into static selection~\cite{moso,moderate,yang2025rl} and dynamic selection~\cite{infobatch,dynamic_pruning,dynamic_pruning-2}.
Static selection identifies a fixed subset of data before training begins, whereas dynamic selection adjusts the data during model training, enabling better adaptation of the training data to the evolving learning process.

While achieving promising results, existing methods face two fundamental limitations.
First, most approaches rely on task- or architecture-specific handcrafted metrics to estimate sample importance, e.g., clustering-based statistics~\cite{yang2025rl,moderate} or gradient-derived scores~\cite{moso,tdds}.
Such criteria are often tailored to image classification, and are difficult to generalize to paradigms with different supervision and optimization structures, such as object detection.
As a result, extending these methods to new tasks typically requires substantial task-specific redesign, hindering scalability and applicability~\cite{infobatch}. 
Second, sample utility is inherently dynamic and evolves throughout training, yet most methods rely on a converged surrogate model or snapshot-based scores for selection~\cite{infobatch,beyond,d2}. As observed in~\cite{moso}, such evaluations potentially favor samples that are difficult or influential in later training stages, and can be effected by transient training fluctuations.
Together, these limitations raise an interesting and pressing question: \textit{Can we design an agent that adaptively selects data on the fly, while scaling across tasks in a plug-and-play manner?}

\begin{figure*}
    \centering
    \includegraphics[width=0.8\linewidth]{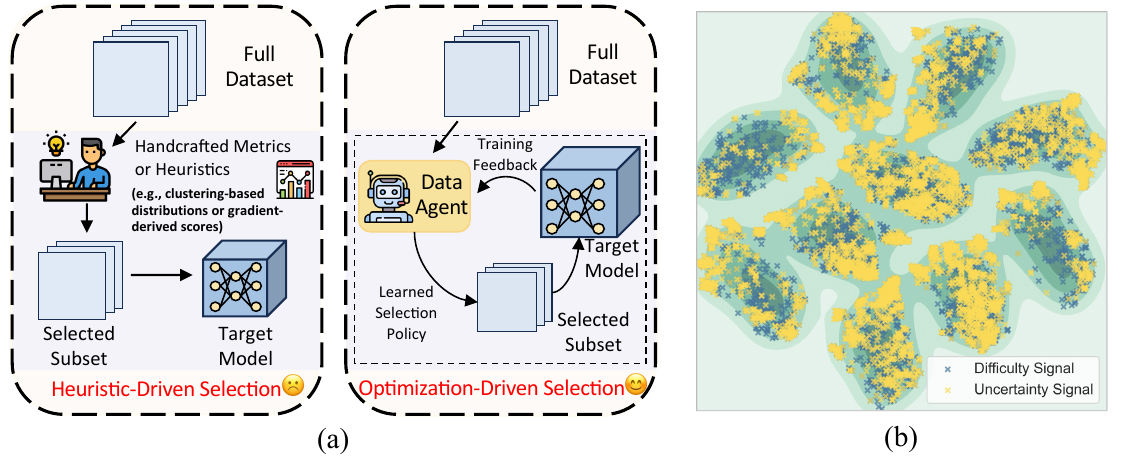}
    \caption{(a) \textbf{End-to-end dynamic data selection.} Existing methods often rely on handcrafted, task-specific static heuristics to estimate sample importance, limiting the scalability across learning paradigms. In contrast, our framework formulates data selection as a learning problem and jointly optimizes it with model training in a plug-and-play manner, forming a closed-loop, training-aware selection process. (b) \textbf{Illustration of data points prioritized by difficulty and uncertainty signals.} The uncertainty signal concentrates on the inter-cluster boundaries and transitional regions, while the difficulty signal focuses more on the sparse cluster areas. }
    \label{fig:fig1}
    \vspace{-3mm}
\end{figure*}
Our answer is a resounding \textit{Yes}!
In this paper, we propose Data Agent, an end-to-end dynamic data selection framework that adaptively selects training data throughout training.
Given that both sample utility and model states evolve throughout training, we formulate dynamic data selection as a sequential decision-making problem, where an agent learns a sample-wise selection policy that co-evolves with model optimization. 
At each training stage, the agent observes the current model state and determines which samples to prioritize.
It is guided by a composite signal that combines a loss-based difficulty measure and a confidence-based uncertainty measure.
Intuitively, difficult samples often correspond to underrepresented patterns in the data distribution, while uncertain samples highlight regions near the decision boundary.
Our theoretical analyses (Prop. \ref{prop:diff}/\ref{prop:infor_gain}) further prove that difficulty and uncertainty signals target complementary objectives, respectively prioritizing samples with larger optimization impact and those yielding higher expected information gains.
To automatically balance these objectives over training, we introduce a tuning-free, self-adaptive reward weighting mechanism. 
Early in training, the agent emphasizes difficult samples to accelerate representation learning, while later it gradually shifts focus toward uncertain samples to refine decision boundaries and improve generalization.
This adaptive process enables data selection to co-evolve with the model's learning dynamics.
Importantly, due to its dataset-agnostic formulation and modular reward design, Data Agent scales effortlessly across various learning paradigms.
It can be seamlessly applied to object detection, semantic segmentation, and LLM instruction tuning, enabling lossless training acceleration in a plug-and-play manner. 
When augmented with cross-modality semantic consistency signals~\cite{yang2024clip}, it further exhibits robustness to noisy and corrupted data, highlighting its applicability in real-world settings.

 
Extensive experiments across a wide range of datasets, architectures, and tasks demonstrate that our method consistently accelerates training while preserving or even improving performance with high efficiency. 
On large-scale ImageNet-1k~\cite{imagenet}, our method reduces training costs by over 50\% while improving performance compared to the full dataset.
Beyond image classification, our framework generalizes across tasks such as object detection, semantic segmentation, and LLM instruction tuning.
This also highlights its strong cross-architecture generalization, including ResNet~\cite{resnet}, ViT~\cite {vit}, YOLO~\cite{yolov8}, UperNet~\cite{upernet}, and LLaMA~\cite{llama}.
Notably, on MMLU~\cite{mmlu}, our method outperforms the full-dataset baseline by 2\% with only 50\% of the data on widely used LLaMA-7B. 
Moreover, even in noisy datasets, our method demonstrates strong robustness, outperforming existing baselines by at least 8\%, underscoring its reliability.
Our main contributions are as follows: 
\textbf{1)} We formulate data selection as a training-aware sequential decision-making problem and propose \textbf{Data Agent}, an end-to-end framework that learns a sample-wise selection policy co-evolving with model training. 
\textbf{2)} We introduce a composite reward integrating sample difficulty and model uncertainty, together with an adaptive reward weighting mechanism, enabling tuning-free optimization.
\textbf{3)} With a dataset-agnostic formulation and modular reward structure, Data Agent scales across tasks, architectures, and scenarios, serving as a plug-and-play module.
\textbf{4)} Experiments demonstrate that our method consistently outperforms SOTA approaches, achieving lossless training acceleration with over 50\% reduction in training cost and saving tens to over one hundred GPU hours across deep models.





\section{Related Work}
Data-efficient learning can be broadly categorized into static data selection~\cite{dataset_pruning,data_diet,yang2024clip,wang2026winningpruninggambleunified}, dynamic data selection~\cite{infobatch,dynamic_pruning,dynamic_pruning-2,yang2025multimodalguideddynamicdatasetpruning,yang2025dynamic}, dataset distillation~\cite{dataset_distillation,dataset_distillation2,cazenavette2025datasetdistillationpretrainedselfsupervised,li2025hyperbolicdatasetdistillation,dataset_distillation3,su2024d}, and dataset condensation~\cite{dataset_condensation1,dataset_condensation2,shao2024elucidating,malakshan2025decomposed}.
Static pruning is efficient and typically one-shot, but cannot adapt to training dynamics. Dynamic pruning captures evolving sample utility during training, at the cost of modest online overhead.
Following dynamic data selection, we propose Data Agent capable of identifying effective training coresets during online training.

\subsection{Static Data Selection}
Static data selection aims to identify a compact yet representative subset of the full datasets before training begins. 
Models trained on such coresets can achieve results comparable to those on the original dataset.
Existing methods are typically based on predefined or heuristic metrics. These metrics can be broadly categorized into importance-criteria-based~\cite{forgetting}, dataset-distribution-based~\cite{ccs}, and optimization-based~\cite{yang2024clip}.
Among importance-criteria-based methods, EL2N and GraNd~\cite{data_diet} calculate the gradient norm and error-$\ell_2$-norm. Forgetting~\cite{forgetting} estimates the samples' misclassification frequency for selection.
MoSo~\cite{moso} estimates the effect of removing each sample from the training set, and Memorization~\cite{score-based-3} assesses the impact of a sample's absence on the model's ability.
DUAL~\cite{dual} leverages difficulty and uncertainty scores to identify important samples from the early training stage with high efficiency.
The work~\cite{yang2024clip} leverages multimodal features to filter out noisy samples by prioritizing semantically aligned samples. 

Based on dataset distribution, Herding~\cite{herding} selects samples closer to the corresponding class centers.
D2~\cite{d2} estimates sample difficulty based on its neighbors' difficulty, and Moderate~\cite{moderate} selects samples with closer distances to the median score.
Moreover, CCS~\cite{ccs} evaluates the dataset coverage by extending the classical set cover problem to the distribution cover problem. RL-Selector~\cite{yang2025rl} estimates sample coverage throughout the entire training process and leverages RL to select a fixed subset, requiring the entire training process for each selection ratio. The work~\cite{ramalingam2023weightedkcenteralgorithmdata} proposes to compute subsets based on the k-center and uncertainty sampling.

Methods based on optimization algorithms optimize the selected datasets based on gradient matching~\cite{opt-based-3,core-set}, self-supervised metrics~\cite{beyond}, influence functions~\cite{dataset_pruning}, bi-level optimization~\cite{glister}, facility location function~\cite{craig,crest}, temporal dual-depth scoring~\cite{tdds}, prediction uncertainty~\cite{dyn-unc}, and submodularity function~\cite{cgscore,opt-based-1,opt-based-4,submodular}. 
Despite the promising results, these works face limitations: 1) The predefined or handcrafted metrics hardly work well across architectures and datasets~\cite{infobatch}; 2) To evaluate the training effect of samples, these methods typically introduce substantial additional costs.

\subsection{Dynamic Data Selection}
Dynamic data selection selects data points during online training, allowing the training data to adapt to the models' training stages.
The work~\cite {dynamic_pruning} proposes UCB and $\epsilon$-greedy algorithms to select samples with the highest uncertainty during training.
The work~\cite{liu2023dataefficientaugmentationtrainingneural} proposes a data-efficient framework that extracts small subsets of training data for augmentation to achieve comparable performance to the full datasets. 
SAS~\cite{data-efficient-contrastive-ssl} observes that the most influential samples for contrastive learning contribute the least to supervised learning, and proposes an algorithm to select subsets that maximize augmentation similarity to the full data.
The work~\cite{dynamic_pruning-2} emphasizes the importance of samples that are used for training a few times and proposes a scoring mechanism based on UCB and $\epsilon$-greedy.
Differently, OPUS~\cite{opus} proposes an optimizer-induced dynamic selection, which formulates data utility through optimizer-induced update dynamics, and GREATS~\cite{greats} optimizes batch quality via Taylor expansion to reduce validation loss.
Recently, InfoBatch~\cite{infobatch} proposes an unbiased dataset pruning method that can accelerate training by pruning less informative samples based on the loss distribution, reducing training costs without degrading performance.
\section{The Proposed Method}
\begin{figure}
    \centering
    \includegraphics[width=1.\linewidth]{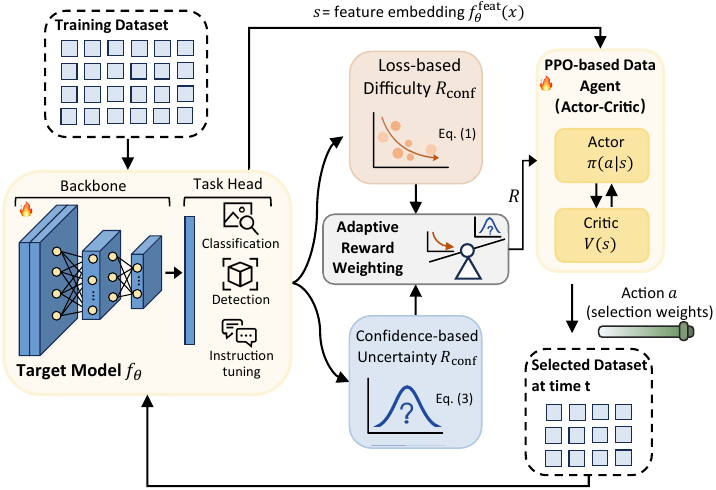}
    \caption{The framework of the proposed Data Agent. At each training stage, the agent observes the model state and derives reward signals from standard forward passes. These signals are combined using an adaptive weighting mechanism to guide a PPO-based actor-critic agent, which learns the selection policy. 
    The selected data is used in subsequent training, forming a closed-loop training pipeline where data selection evolves alongside model optimization. Notably, the modular reward design enables the framework to be easily adapted to various learning paradigms.    
    }
    \label{fig:framework}
    \vspace{-2mm}
\end{figure}
\noindent \textbf{Overview of the Data Agent.}
Dynamic data selection is inherently a sequential decision-making problem that evolves alongside model training, where the utility of data samples changes as the model learns.
Here, we propose the Data Agent, a lightweight PPO-based framework that adaptively determines the training data distribution throughout training.
At each training stage, the data agent observes the current target model state and optimizes a sample-wise policy, guided by two complementary, training-aware signals: a loss-based difficulty signal and a confidence-based uncertainty signal.
These signals are directly derived from the forward passes of the target model, capturing the immediate contribution of a sample to empirical risk minimization and the model's predictive uncertainty, respectively.
To enable an end-to-end optimization, we introduce an adaptive reward weighting mechanism that automatically adjusts the relative importance of these signals throughout training.
This allows the data agent to focus on difficult signals early on to accelerate representation learning, while shifting attention toward uncertainty as training progresses to refine decision boundaries and improve generalization.
Notably, due to the dataset-agnostic framework and modular reward structure, our method is highly scalable across diverse tasks, e.g., object detection, and scenarios, e.g., noisy datasets.


\subsection{Reinforcement Learning Formulation of Data Selection}
\noindent \textbf{Preliminary.} We formulate dynamic data selection problem as a Markov decision process (MDP)~\cite{zhong2024dpo}, denoted by the tuple $\mathcal{M}=(\mathcal{S}, \mathcal{A}, \mathcal{P}, r, \rho, H)$, where $\mathcal{S}$ is the state space, $\mathcal{A}$ the action space, $\mathcal{P}$ the transition kernel, $r$ the reward function, $\rho$ the initial state distribution, and $H$ the horizon length.
A policy $\pi(a|s)$ defines a distribution over actions conditioned on the current state $s$.
At each step, the agent observes the current state $s$ induced by the target model, updates the policy, and receives rewards derived from the training-aware signals.
The agent's goal is to optimize a policy that identifies the most informative training samples for the model.

\noindent \textbf{State Space.}
The state space is defined by the internal representations of the target model, which capture the current training state for each sample.
Let $f_\theta(\cdot)$ denote the backbone network parameterized by $\theta$.
For a given sample $x$, the state is defined as: $s=f_\theta^{\text{feat}}(x)$, where $f_\theta^{\text{feat}}(\cdot)$ is the feature embedding output from the backbone network.
For example, in image classification tasks, this corresponds to the outputs before the final fully connected layer.
As training progresses, both the feature space and the model's representation structure change continuously.
Thus, the state not only encodes sample-specific information but also captures the model's progress, enabling the data agent to condition its selection policy on both the inherent properties of each sample and the evolving training dynamics.

\noindent \textbf{Action Space.}
The action space is designed for adaptive control over the data distribution.
Rather than selecting or discarding samples via discrete decisions, the agent outputs a continuous-valued action for each sample, which avoids the combinatorial complexity and non-differentiability inherent to subset selection.
Given a state $s$, the policy $\pi(a|s)$ produces an action $a\in [0,1]$, representing the selection weight for each sample.
This continuous formulation transforms the dynamic data selection problem into a differentiable control problem, which allows for stable and efficient policy optimization in a non-stationary training environment.

\noindent \textbf{Training-aware Reward Design.}
The reward function guides the agent toward adaptive data selection and is computed from training-time forward passes, without relying on a validation set.
Importantly, sample utility evolves with model learning: samples that are informative at early training stages may become redundant as the model's representations improve.
Thus, we propose a training-aware composite reward that models the interplay between data and the evolving model via two complementary signals: sample difficulty and model uncertainty.
This combination encourages focusing on samples that are crucial for optimization and rich in information at the current training stage. 

\noindent \textit{Loss-based Difficulty Reward:} 
We introduce a loss-based difficulty reward to estimate the immediate learning difficulty of each sample.
Specifically, given the per-sample training loss $\mathcal{L}$ and target model $f_\theta$, the difficulty reward is defined as:
\begin{equation}\label{eq:r_diff}
    R_{\text {diff}}(x_i,y_i) = \mathcal{L}(f_\theta(x_i), y_i).
\end{equation}
This reward has three key advantages~\cite{infobatch,cilimkovic2015neural}: 1) loss values can be directly obtained from standard forward passes, avoiding additional computational overhead; 2) it naturally adapts to the evolving training stage of $f_\theta$; and 3) it is task- and architecture-agnostic, making it broadly applicable to various learning paradigms.
\begin{proposition}\label{prop:diff}
    Let $f_\theta$ be a network trained with the CE loss $\ell(x,y)$ and softmax outputs $p_\theta(y|x)$.
Under a first-order SGD update, the expected magnitude of the parameter update induced by a sample $(x_i,y_i)$ satisfies
\begin{equation}
    \|\nabla_\theta \ell(x_i,y_i)\| \;\propto\; 1 - p_\theta(y_i \mid x_i),
\end{equation}
and is therefore a monotonic function of the training loss $\ell(x_i,y_i) = -\log p_\theta(y_i \mid x_i)$.
\end{proposition}
Proposition~\ref{prop:diff}, proved in Appendix~\ref{sec:appendix-proof-prop0}, shows that prioritizing samples with higher loss accelerates empirical risk minimization by emphasizing those samples with larger optimization impact.


\noindent \textit{Confidence-based Uncertainty Reward:}
While the difficulty reward captures optimization pressure, it is limited to the likelihood of the annotated class and neglects the uncertainty of the model's prediction.
For instance, a loss-based strategy may overlook samples that are correctly classified but still uncertain, particularly those near decision boundaries. 
These samples are critical for refining the decision boundaries.
To address this, we introduce a confidence-based uncertainty reward, derived from the predictive entropy:
\begin{equation}\label{eq:r_conf}
    R_{\text {conf}} = -\sum_{c=1}^C p_\theta\left(y=c \mid x_i\right) \log p_\theta\left(y=c \mid x_i\right),
\end{equation}
where $C$ is the number of classes.
This reward can be obtained via standard forward passes without additional computation cost.
By favoring samples with high predictive uncertainty, the data agent is encouraged to focus on boundary-sensitive samples, which are essential for improving decision reliability and generalization.

\begin{proposition}\label{prop:infor_gain}
    Let $p_\theta(y|x)$ denote the predicted distribution. Under a first-order SGD update on the CE loss, the expected information gain of learning a sample $x_i$ is proportional to the predictive entropy:
    \begin{equation}
        \mathbb{E}_{y \sim p_\theta(y \mid x)}\left[D_{\mathrm{KL}}\left(p_{\theta^{\prime}}(\cdot \mid x) \| p_\theta(\cdot \mid x)\right)\right] \propto H\left[p_\theta(y \mid x)\right],
    \end{equation}
    where $\theta^{\prime}=\theta-\eta \nabla_\theta \ell(x, y)$,
\end{proposition}
where $D_{\mathrm{KL}}$ and $H$ are KL-divergence and predictive entropy, respectively.
Proposition~\ref{prop:infor_gain}, proved in Appendix~\ref{sec:appendix-proof-prop1}, shows that prioritizing uncertain samples approximately maximizes the information gain for model learning.
Taken together, Proposition~\ref{prop:diff} and ~\ref{prop:infor_gain} highlight the complementary nature of the difficulty and uncertainty rewards, which capture different but crucial aspects of sample utility.

\noindent \textbf{Adaptive Reward Weighting.} 
The relative importance of difficulty and uncertainty signals varies across different training stages.
In early stages, when model representations are still forming, the difficulty signal drives representation learning by prioritizing challenging samples.
As training progresses, the uncertainty in the model's predictions becomes more informative, enabling finer decision refinement.
To capture this dynamic shift and avoid manually tuning, we propose an adaptive reward weighting mechanism that automatically adjusts the contributions of the difficulty and uncertainty rewards based on training dynamics.
Since the variance of each reward signal reflects its informativeness~\cite{tdds,chen2018gradnormgradientnormalizationadaptive}, we compute the weighting coefficient $r$ as follows:
\begin{equation}\label{eq:weighting}
    r = \frac{Var(R_{\text {diff}})}{Var(R_{\text {diff}})+Var(R_{\text {conf}})+\epsilon},
\end{equation}
where $\epsilon$ is a small constant for numerical stability.
The final reward is computed as:
\begin{equation}\label{eq:reward}
    R=r \cdot R_{\text {diff}}+(1-r) \cdot R_{\text {conf}}.
\end{equation}
This formulation allows the agent to adjust its selection focus in an end-to-end, data-driven manner without the need for external hyperparameter tuning.
It encourages the agent to balance reducing empirical risk and maximizing epistemic information gain, ultimately enhancing model generalization.
Notably, the reward design is modular and extensible, allowing additional task-specific or scenario-specific signals to be adjusted or incorporated as needed, further enhancing the framework's flexibility. 


\begin{table*}[]
    \centering
    \caption{Comparison with state-of-the-art baselines. All methods are trained using ResNet-18 on CIFAR-10/100 and ResNet-50 on Tiny-ImageNet. Random* refers to randomly selecting samples in each epoch. Some results are from~\cite{infobatch}. \label{tab:comparison_experiment}}
	\resizebox{0.75\textwidth}{!}{
    \begin{tabular}{c|ccc|ccc|ccc}
    \bottomrule[1.5pt]
    Dataset &  \multicolumn{3}{c|}{\cellcolor{mygray}CIFAR-10}& \multicolumn{3}{c|}{\cellcolor{mygray}CIFAR-100} & \multicolumn{3}{c}{\cellcolor{mygray}Tiny-ImageNet} \\ \hline
    Whole Dataset &\multicolumn{3}{c|}{95.6}&\multicolumn{3}{c|}{78.2} &\multicolumn{3}{c}{45.0} \\ \hline
    Selection Ratio (\%)& 30& 50&70& 30& 50&70& 30& 50&70 \\ \hline
     \multicolumn{10}{c}{\textbf{Static selection methods}} \\ \hline
    Random &90.2&92.3&93.9&69.7&72.1&73.8&29.8&37.2&42.2 \\ 
    EL2N~\cite{data_diet} &91.6&95.0&95.2 &69.5&72.1&77.2 &26.6&37.1&44.0 \\
    GraNd~\cite{data_diet} &91.2&94.6&95.3 &68.8&71.4&74.6 &29.7&36.3&43.2 \\
    Forgetting~\cite{forgetting} &91.7&94.1&94.7 &69.9&73.1&75.3 &28.7&33.0&41.2 \\
    RL-Selector~\cite{yang2025rl} &91.8&-&95.4 &71.1&-&77.6 &31.1&-&44.5 \\
    Herding~\cite{herding} &80.1 &88.0&92.2 &69.6&71.8&73.1&29.4&31.6&39.8 \\
    Moderate~\cite{moderate} &91.5&94.1&95.2&70.2&73.4&77.3 &30.6&38.2&42.8 \\
    Glister~\cite{glister} &90.9&94.0&95.2 &70.4&73.2&76.6 &30.1&39.5&43.9 \\
    DP~\cite{dataset_pruning}&90.8&93.8&94.9 &-&73.1&77.2 &-&-&- \\
    Self-sup. prototypes~\cite{beyond} &91.0&94.0&95.2&70.0&71.7&76.8 &27.7&37.9&43.4 \\
    MoSo~\cite{moso} &91.1&94.2&95.3 &70.9&73.6&77.5 &31.2&38.5&43.4 \\
    CLIP-Sel~\cite{yang2024clip} &91.9&94.5&95.0 &70.8&73.7&77.0 &31.7&40.0 &46.0 \\ \hline
    \multicolumn{10}{c}{ \textbf{Dynamic pruning methods}} \\ \hline
    Random* &93.0&94.5&94.8 &74.4&75.3&77.3 &41.5&42.8&43.1 \\
    UCB~\cite{dynamic_pruning} &93.9&94.7&95.3 &-&75.3&77.3 &-&-&-    \\
    $\epsilon$-Greedy~\cite{dynamic_pruning} &94.1&94.9&95.2 &-&74.8&76.4&-&-&- \\
    InfoBatch~\cite{infobatch} &94.7&95.1&95.6 &76.5&78.1&78.2 &42.2&43.2&43.8 \\ 
    \hline
    Ours &\textbf{95.0}&\textbf{95.3}&\textbf{96.0}&\textbf{77.6}&\textbf{78.9}&\textbf{79.5}&\textbf{44.9}&\textbf{47.0}&\textbf{49.4} \\
    \bottomrule[1.5pt]
    \end{tabular}}
    \vspace{-2mm}
\end{table*}

\noindent \textbf{PPO-based Agent Optimization.}
We optimize a sample-wise selection policy using Proximal Policy Optimization (PPO), which dynamically controls the training data distribution during model training.
Specifically, the actor and critic networks are parameterized as three-layer MLPs, denoted as $\theta_\pi$ and $\theta_v$, respectively.
The policy determines the sample-wise selection weights.
Samples with the top-$k$ highest action weights are selected.
As the target model and reward signals evolve during training, unconstrained updates to the policy can lead to abrupt changes in the data selection process, potentially destabilizing the joint optimization of the model and the data agent.
To mitigate this, we adopt PPO to constrain policy updates and stabilize the co-evolutionary process.
PPO’s clipped objective enables stable, incremental updates to the policy, while retaining the advantages of policy gradient methods:
\begin{equation}\label{eq:loss_actor}
\begin{aligned}
    \mathcal{L}_{\text{actor}}(\theta_\pi) = & \mathbb{E}_t[\min(\omega_t(\theta_\pi)\hat{A}_t, \\
    & \text{clip}\sbr{\omega_t(\theta_\pi), 1-\epsilon, 1+\epsilon}\hat{A}_t)],
\end{aligned}
\end{equation}
where $\omega_t(\theta_\pi)$ is the probability ratio between the current and previous policies, $\hat{A}_t$ is the advantage estimate at time $t$, and $\epsilon$ controls the range of clipping.
In addition, we compute the advantage function using generalized advantage estimation (GAE)~\cite{schulman2018highdimensionalcontinuouscontrolusing}, which balances bias and variance in non-stationary environments.
The temporal-difference residual $\delta_t$ at time $t$ is
\begin{equation}
\delta_t = r_t + \gamma V(s_{t+1}) - V(s_t),
\end{equation}
where the advantage estimate $\hat{A}_t$ is computed as:
\begin{equation}
\hat{A}_t = \sum_{l=0}^{T-t-1} (\gamma \lambda)^l \delta_{t+l},
\end{equation}
where $\lambda$ is a parameter that controls the trade-off between bias and variance.
The critic network is trained to approximate the state-value function $V(s)$ by minimizing the squared temporal-difference error.
Given the advantage $\hat{A}_t$ computed by GAE, the value function loss is then given by
\begin{equation}\label{eq:loss_critic}
\mathcal{L}_{\text{critic}}(\theta_v)
= \mathbb{E}_t \mbr{\sbr{V_{\theta_v}(s_t) - \hat{A}_t - V(s_t)}^2}.
\end{equation}

\section{Experiment}
\subsection{Experiment Setup}
\noindent \textbf{Datasets and Network Architectures.}
We evaluate our framework across diverse benchmarks spanning a wide variety of tasks to demonstrate its generalization and scalability.
For image classification, we use both coarse-grained and fine-grained benchmarks, including CIFAR-10/100~\cite{cifar100}, Tiny-ImageNet~\cite{tiny}, and ImageNet-1k~\cite{imagenet}.
We further extend the evaluation to semantic segmentation on ADE20k~\cite{ade20k}, object detection on MS-COCO~\cite{mscoco}, and LLM instruction tuning on MMLU~\cite{mmlu} and AlpacaEval 2.0~\cite{alpacaeval}.
Moreover, we evaluate the generalization of our framework to more challenging scenarios, including ImageNet-O~\cite{imagenet-a}, ImageNet-Hard~\cite{imagenet-hard}, and ImageNet-R~\cite{imagenet-r}.
To assess its cross-architecture generalization, we leverage a range of deep models, including ResNet series~\cite{resnet}, and ViT series~\cite{vit} for classification, YOLOv8~\cite{yolov8} for detection, UperNet~\cite{upernet} for semantic segmentation, and LLaMA 7B~\cite{llama} for LLM instruction tuning.

\begin{table*}[]
       \centering
    \caption{Results on ImageNet-1k with a 60\% selection ratio using ResNet-50 on an 8-A100 server. We report wall-clock time (h) and total GPU time (GPU hours). Note that, due to high computational and memory costs~\cite{moderate}, Glister and CG-Score are not reported. Some results are from~\cite{infobatch}.}\label{tab:imagenet-1k}
	\resizebox{0.95\textwidth}{!}{
    \begin{tabular}{c|ccccccccccccc}
    \toprule[1.pt]
    Method &Herding&EL2N&GraNd&Forgetting&RL-Selector&SSP&Moderate&CLIP-Sel&UCB&Infobatch &Ours&Whole Dataset \\ \hline
    Acc. (\%) &71.1&72.3&71.0&72.5& 73.4&70.0&73.1&73.2&75.8&76.5 &\textbf{76.8}&76.4 \\ \hline
    Time (h) &10.5&10.5&10.5&10.5&10.5&10.5&10.5&10.5&10.5& 10.5&10.5&17.5 \\
    Overhead (h) &$>$17.5&$>$17.5&$>$17.5&$>$17.5&$>$17.5&$>$24.0&$>$17.5&$>$1.6&0.03&0.0028& 0.125 &0.0 \\
    Overall (n*h) & $>$224.0 &$>$224.0&$>$224.0&$>$224.0&$>$224.0&$>$276.0&$>$224.0&$>$96.8&84.0& 84.0& 85.0 &140.0 \\
    \bottomrule[1.pt]
    \end{tabular}}
\end{table*}
\begin{figure*}
\centering
    \begin{minipage}{0.57\textwidth}
        \centering
    \includegraphics[width=1\textwidth]{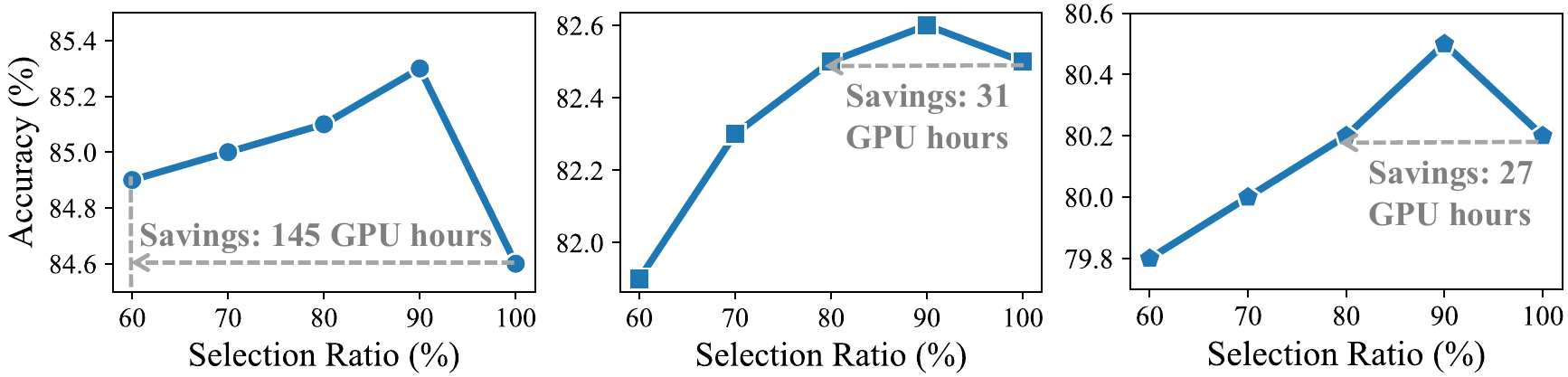}
    \caption{Performance and saved costs on ImageNet-1k across Swin-T, ViT-B, and ViT-L on a 4-A100-GPU server. We report the total GPU hours. \label{fig:vit-imagenet} }
    \end{minipage}
    \hspace{3mm}
    \begin{minipage}{0.4\textwidth}
        \centering
        \captionof{table}{Generalization to LLM instruction tuning using LLaMA-7B with a 50\% selection ratio. }
        \setlength{\tabcolsep}{9pt}
	\resizebox{1\textwidth}{!}{
		\begin{tabular}{c|c|cc}
			\toprule[1.pt]
   & \multirow{2}{*}{MMLU} & \multicolumn{2}{c}{AlpacaEval 2.0} \\ \cline{3-4}
   &&Win Rate & LC Win Rate \\ \hline
   Full Dataset & 34.9& 1.9 & 6.7 \\
   Random* & 34.6 & 1.7 & 6.0 \\
   Ours & \textbf{36.9} & \textbf{2.0} & \textbf{7.7} \\
			\bottomrule[1.pt]
		\end{tabular}
	}
	\label{tab:llm-ft}
    \end{minipage}
    \vspace{-2mm}
\end{figure*}

\noindent \textbf{Comparison with State-of-the-arts.} We compare our method with a wide range of static and dynamic data selection methods, including 1) Random, 2) EL2N~\cite{data_diet}, 3) GraNd~\cite{data_diet}, 4) Forgetting~\cite{forgetting}, 5) RL-Selector~\cite{yang2025rl}, 6) Herding~\cite{herding}, 7) Moderate~\cite{moderate}, 8) Glister~\cite{glister}, 
9) DP~\cite{dataset_pruning}, 10) MoSo~\cite{moso}, 11) CLIP-Selector~\cite{yang2024clip}, 12) Self-sup. prototypes~\cite{beyond}, 13) Random*, 14) InfoBatch~\cite{infobatch}, 15) $\epsilon$-Greedy~\cite{dynamic_pruning-2}, and 16) UCB~\cite{dynamic_pruning-2}.

\subsection{Performance Comparison}
\noindent \textbf{Results on CIFAR-10/100 and Tiny-ImageNet.} As shown in Table~\ref{tab:comparison_experiment}, we compare our proposed method with both static and dynamic data selection methods on CIFAR-10/100 and Tiny-ImageNet. 
Our method consistently outperforms existing methods by a large margin, achieving significant training acceleration without sacrificing model performance.
Specifically, it achieves comparable or better accuracy compared to the full dataset with only 50\% of the data on CIFAR-100 and 30\% on Tiny-ImageNet.
While performance tends to decrease as the selection ratio decreases, our method shows the least drop in accuracy compared to full-dataset training.
These results underscore the effectiveness of our adaptive, training-aware selection policy, which dynamically adjusts data selection throughout training to maintain high model performance with fewer data.

\noindent \textbf{Results on ImageNet-1k.} 
As shown in Table~\ref{tab:imagenet-1k}, in addition to the performance comparison, we also compare computation efficiency on ImageNet-1k with a 60\% selection ratio.
Thus, we report the training time, the introduced overhead, and the total GPU hours for various methods.
Notably, our method achieves a 0.4\% accuracy improvement over the full-dataset baseline while reducing training costs by nearly 40\%, saving over 55 GPU hours.
Importantly, since most static methods require training a surrogate model to estimate sample importance, the computational overhead is higher.
Thus, our method outperforms static selection methods in both accuracy and efficiency.
When compared to other dynamic selection methods, our approach delivers superior performance while maintaining competitive computational efficiency. 
This efficiency is attributed to our lightweight PPO-based data agent, which operates directly on model features using a compact architecture consisting of just three linear layers, minimizing computation.
Meanwhile, the reward can be directly obtained via standard forward passes without introducing complex algorithms.
Overall, these results demonstrate that our method delivers an effective and scalable solution on large-scale datasets.

\subsection{Generalization to More Advanced Architectures}
To further evaluate the cross-architecture generalization, we apply the data agent to the training of ViT-based models, including ViT-Base, ViT-Large, and Swin-Transformer.
As shown in Figure~\ref{fig:vit-imagenet}, our method achieves significant training acceleration, with no loss in performance, across different selection ratios.
For instance, on ViT-L, our method reduces overall training time by more than 150 GPU hours using only 60\% of the data, without sacrificing accuracy.
Notably, at higher selection ratios, our method achieves higher accuracy, outperforming full-dataset training.
These results highlight the architecture-agnostic nature of our method, demonstrating its scalability.


\begin{table}[]
    \centering
        \caption{Object detection mAP (\%) on MS-COCO using YOLOv8~\cite{yolov8}. }
    \setlength{\tabcolsep}{9pt}
	\resizebox{0.45\textwidth}{!}{
		\begin{tabular}{c|ccc|c}
			\toprule[1.pt]
   \multicolumn{1}{c|}{Selection Ratio}& 70\% & 80\% & 90\% & 100\% \\ \hline
    Random* & 37.5 & 37.9 & 38.5& \multirow{2}{*}{39.6}  \\ 
	Ours		& \textbf{38.5}& \textbf{39.0} & \textbf{39.6} &   \\
			\bottomrule[1.pt]
		\end{tabular}
	}
	\label{tab:detection}
    \vspace{-2mm}
\end{table}
\subsection{Generalization to Different Training Paradigms}

\begin{table}[]
    \centering
        \caption{Segmentation mIoU (\%) on ADE20K using UperNet~\cite{upernet}. }
        \setlength{\tabcolsep}{9pt}
	\resizebox{0.45\textwidth}{!}{
		\begin{tabular}{c|ccc|c}
			\toprule[1.pt]
   \multicolumn{1}{c|}{Selection Ratio}& 70\% & 80\% & 90\% & 100\% \\ \hline
   Random* &45.1&45.3&45.7&\multirow{2}{*}{45.4} \\
		Ours	&\textbf{46.4} & \textbf{46.5} & \textbf{46.6} &  \\
			\bottomrule[1.pt]
		\end{tabular}
	}
	\label{tab:segmentation}
    \vspace{-5mm}
\end{table}
\noindent \textbf{Object Detection.}
While most existing data selection approaches cannot be extended to detection tasks, we demonstrate the versatility of our data agent by integrating it into object detection pipelines.
Specifically, we apply it to YOLOv8 on MS-COCO, where the sample-wise detection loss naturally serves as the difficulty-based reward signal, without relying on the confidence reward.
As shown in Table~\ref{tab:detection}, our data agent achieves lossless or comparable performance when trained with only 70-90\% of the data.
These results show that our method is not tied to classification-specific supervision or architectures, but can also adapt to dense prediction settings, highlighting its strong scalability and broad applicability.

\noindent \textbf{Semantic Segmentation.}
In addition to object detection, we further assess the generality of our proposed framework by applying it to semantic segmentation, using UperNet on ADE20K.
As shown in Table~\ref{tab:segmentation}, our method not only preserves segmentation performance but also achieves improvements when trained with only 70-90\% of the full dataset.
These results demonstrate that our method can adapt to dense supervision tasks, highlighting its scalability and architecture-agnostic nature.


\noindent \textbf{LLM Instruction Tuning.}
Beyond vision tasks, our data agent also accelerates LLM instruction fine-tuning, particularly with LLaMA-7B across MMLU and AlpacaEval 2.0.
MMLU evaluates multi-domain language understanding, while AlpacaEval 2.0 assesses instruction-following and alignment quality.
In this context, the agent dynamically selects informative instruction-response pairs during fine-tuning.
As shown in Table~\ref{tab:llm-ft}, even with only 50\% of the training data, our method achieves consistent improvements across all metrics.
These results validate the effectiveness of our approach for LLM training pipelines, and underscore the framework's modality- and task-agnostic scalability for accelerating large-scale model training.


\subsection{Generalization under Distribution Shift}
\begin{table}[]
    \centering
        \caption{Generalization performance of ResNet-50 trained with our method on ImageNet-Hard/R/O. We report AUPR (\%) on ImageNet-O and accuracy (\%) on others. }
        \setlength{\tabcolsep}{9pt}
	\resizebox{0.45\textwidth}{!}{
		\begin{tabular}{c|cccc|c}
			\toprule[1.pt]
   \multicolumn{1}{c|}{Selection Ratio}&60\% & 70\% & 80\% & 90\% & 100\% \\ \hline
        ImageNet-Hard &14.7&15.3&15.6&15.7& 14.6 \\
        ImageNet-R &38.0&38.2&38.5&38.7& 36.2 \\
        ImageNet-O &15.7&15.6&15.7&15.8& 13.2 \\
			\bottomrule[1.pt]
		\end{tabular}
	}
	\label{tab:ood-benchmark}
\end{table}
To further evaluate the generalization of our method, we conduct experiments on several challenging out-of-distribution benchmarks, i.e., ImageNet-O/R/Hard.
These datasets are designed to assess a model's ability to generalize to unseen, more difficult distributions.
We compare models trained on the full ImageNet-1k dataset with those trained using our dynamically selected subsets.
As shown in Table~\ref{tab:ood-benchmark}, our method significantly improves performance across all three benchmarks when trained with only 60-90\% of the data.
These results suggest that our data agent helps the model focus on more informative and representative samples, helping the model learn more generalizable features that are less prone to overfitting specific data distributions.
Importantly, the performance improvements come with reduced training costs, enhancing practicality.


\subsection{Robustness to Noisy Scenarios}
\begin{table}[]
    \centering
    \caption{Robustness to noisy and corrupted data on Tiny-ImageNet using ResNet-50. The noisy ratio is 20\%, and the selection ratios are 20\% and 30\%. Some results are from~\cite{yang2024clip}. \label{tab:noisy-dataset}}
    \resizebox{0.36\textwidth}{!}{
    \begin{tabular}{c|cc|cc}
        \bottomrule[1.2pt]
      \multirow{2}*{Method}   &  \multicolumn{2}{c|}{ Noisy Dataset} &  \multicolumn{2}{c}{ Corrupted Dataset} \\ \cline{2-5}
          & 20\% & 30\% & 20\% & 30\% \\ \hline
        Random & 17.8 & 23.9 & 20.0 & 25.9 \\ 
        Random* &32.5 & 36.1 & 36.5 & 38.3 \\
        CLIP-Sel &26.1 & 33.1 & 26.1 & 32.1 \\ 
        Ours & \textbf{38.1} & \textbf{41.9} & \textbf{41.9} & \textbf{46.7} \\
        \bottomrule[1.2pt]
    \end{tabular}}
    \vspace{-1mm}
\end{table}
Due to the modular design, the proposed data agent can easily incorporate additional signals tailored to specific problem settings.
In real-world scenarios, datasets often contain mislabeled or corrupted images~\cite{moderate}, which can significantly degrade model performance.
To address this, we extend the flexibility of our method by integrating a cross-modality semantic alignment signal~\cite{yang2024clip} to assess the consistency between visual inputs and textual labels.
As shown in Table~\ref{tab:noisy-dataset}, our method consistently outperforms existing SOTA baselines, achieving over 8\% accuracy improvements under noisy label conditions. 
Notably, such robustness emerges naturally from the unified selection mechanism without any modification to the core learning framework or optimization procedure.
These results highlight the adaptability of our method.


\subsection{Ablation Study}
\noindent \textbf{Effect of Different Components.}
As shown in Table~\ref{tab:effect-ablation}, we analyze the effect of the difficulty reward $R_{\text{diff}}$, the confidence reward $R_{\text{conf}}$, and the reward weighting $r$.
Without using any of these, our method, in the first row, degrades to random selection, which yields the lowest accuracy.
This confirms that the proposed reward signals provide a meaningful training-aware supervision signal.
Using $R_{\text{diff}}$ or $R_{\text{conf}}$ alone already improves performance, and combining them is consistently better, suggesting that difficulty and uncertainty capture complementary aspects of sample utility.
Moreover, using $R_{\text{diff}}$ alone yields slightly lower accuracy, as prioritizing difficulty can lead to the selection of ambiguous or unstable samples.
Introducing $r$ further boosts accuracy by dynamically adjusting their relative emphasis over training,  effectively creating a self-adjusting curriculum that evolves with the model and improves generalization.
Thus, removing any component from the framework degrades performance.

\begin{table}[]
    \centering
    \caption{Effect of the difficulty rewards $R_{\text{diff}}$, uncertainty $R_{\text{conf}}$, and adaptive reward weighting $r$ on T-ImageNet using R-50. \label{tab:effect-ablation}}
    \resizebox{0.33\textwidth}{!}{
    \begin{tabular}{ccc|ccc}
    \bottomrule[1.2pt]
   $R_{\text{diff}}$ & $R_{\text{conf}}$ & $r$   &30\%&50\%&70\% \\ \hline
   & & &41.5 & 42.8 & 43.1 \\
   \checkmark& &&42.1&45.0 & 48.2 \\
   &\checkmark&&42.2&45.8 &  48.4 \\
   \checkmark & \checkmark& &42.5 &46.1  &  48.6 \\
   \checkmark &\checkmark& \checkmark &\textbf{44.9}&\textbf{47.0}  &\textbf{49.4} \\
      \bottomrule[1.2pt]
    \end{tabular}}
    \vspace{-2mm}
\end{table}


\begin{figure}
    \begin{minipage}{0.5\textwidth}
        \begin{subfigure}[c]{0.49\textwidth}
		\centering
		\includegraphics[width=4.2cm]{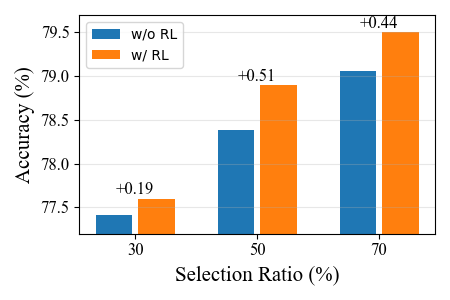}
             \captionsetup{justification=centering}
		\caption{CIFAR-100.}
		\label{fig3-1}
	\end{subfigure}
	\begin{subfigure}[c]{0.49\textwidth}
		\centering
		\includegraphics[width=4.2cm]{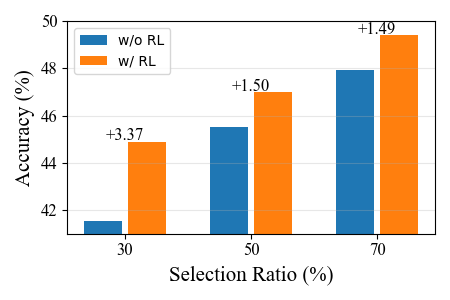}
             \captionsetup{justification=centering}
		\caption{Tiny-ImageNet.}
		\label{fig3-2}
	\end{subfigure}
	\caption{Effect of the RL agent on CIFAR-100 and Tiny-ImageNet under different selection ratios.}
	\label{fig-ablation-RL}
    \end{minipage}
\end{figure}

\noindent \textbf{Effect of Agent Optimization.}
To evaluate the effect of the PPO-based agent, we directly use the composite reward in Eq.~\ref{eq:reward} to select the highest-scoring samples each epoch.
As shown in Figure~\ref{fig-ablation-RL}, incorporating the PPO agent results in consistent accuracy gains across all selection ratios, validating its ability to capture training dynamics and learn a coherent, training-aware selection policy. 
 
\section{Conclusion}
We propose Data Agent, an end-to-end dynamic data selection framework that adaptively selects training data throughout training, substantially accelerating training without sacrificing performance.
Across diverse datasets, architectures, and scenarios, Data Agent achieves enhanced data-efficient learning, cross-task/architecture generalization, and robustness to noisy or corrupted datasets as a plug-and-play module.
We hope this work inspires further research on data-efficient learning and believe Data Agent can serve as a valuable tool for the community, helping reduce computational costs and broaden access to training strong models with limited resources.

\bibliography{example_paper}

@String(CVPR= {IEEE Conf. Comput. Vis. Pattern Recog.})

@String(ECCV= {Eur. Conf. Comput. Vis.})

@String(AAAI = {AAAI})

@String(CVPR  = {CVPR})

@String(ECCV  = {ECCV})

@article{yang2025dynamic,
  title={When Dynamic Data Selection Meets Data Augmentation},
  author={Yang, Suorong and Ye, Peng and Shen, Furao and Zhou, Dongzhan},
  journal={arXiv preprint arXiv:2505.03809},
  year={2025}
}

@article{vit,
  title={An image is worth 16x16 words: Transformers for image recognition at scale},
  author={Dosovitskiy, Alexey and Beyer, Lucas and Kolesnikov, Alexander and Weissenborn, Dirk and Zhai, Xiaohua and Unterthiner, Thomas and Dehghani, Mostafa and Minderer, Matthias and Heigold, Georg and Gelly, Sylvain and others},
  journal={arXiv preprint arXiv:2010.11929},
  year={2020}
}

@article{infobatch,
  title={Infobatch: Lossless training speed up by unbiased dynamic data pruning},
  author={Qin, Ziheng and Wang, Kai and Zheng, Zangwei and Gu, Jianyang and Peng, Xiangyu and Xu, Zhaopan and Zhou, Daquan and Shang, Lei and Sun, Baigui and Xie, Xuansong and others},
  journal={arXiv preprint arXiv:2303.04947},
  year={2023}
}

@inproceedings{tdds,
  title={Spanning training progress: Temporal dual-depth scoring (tdds) for enhanced dataset pruning},
  author={Zhang, Xin and Du, Jiawei and Li, Yunsong and Xie, Weiying and Zhou, Joey Tianyi},
  booktitle={Proceedings of the IEEE/CVF Conference on Computer Vision and Pattern Recognition},
  pages={26223--26232},
  year={2024}
}

@article{dynamic_pruning,
  title={Diversified Batch Selection for Training Acceleration},
  author={Hong, Feng and Lyu, Yueming and Yao, Jiangchao and Zhang, Ya and Tsang, Ivor W and Wang, Yanfeng},
  journal={arXiv preprint arXiv:2406.04872},
  year={2024}
}

@article{dynamic_pruning-2,
  title={Accelerating deep learning with dynamic data pruning},
  author={Raju, Ravi S and Daruwalla, Kyle and Lipasti, Mikko},
  journal={arXiv preprint arXiv:2111.12621},
  year={2021}
}

@inproceedings{
dataset_pruning,
title={Dataset Pruning: Reducing Training Data by Examining Generalization Influence},
author={Shuo Yang and Zeke Xie and Hanyu Peng and Min Xu and Mingming Sun and Ping Li},
booktitle={The Eleventh International Conference on Learning Representations },
year={2023},
url={https://openreview.net/forum?id=4wZiAXD29TQ}
}

@inproceedings{
moderate,
title={Moderate Coreset: A Universal Method of Data Selection for Real-world Data-efficient Deep Learning},
author={Xiaobo Xia and Jiale Liu and Jun Yu and Xu Shen and Bo Han and Tongliang Liu},
booktitle={The Eleventh International Conference on Learning Representations },
year={2023},
}

@inproceedings{
ccs,
title={Coverage-centric Coreset Selection for High Pruning Rates},
author={Haizhong Zheng and Rui Liu and Fan Lai and Atul Prakash},
booktitle={The Eleventh International Conference on Learning Representations },
year={2023},
}

@inproceedings{imagenet,
  title={Imagenet: A large-scale hierarchical image database},
  author={Deng, Jia and Dong, Wei and Socher, Richard and Li, Li-Jia and Li, Kai and Fei-Fei, Li},
  booktitle={2009 IEEE conference on computer vision and pattern recognition},
  pages={248--255},
  year={2009},
  organization={Ieee}
}

@inproceedings{herding,
	title={Herding dynamical weights to learn},
	author={Welling, Max},
	booktitle={Proceedings of the 26th Annual International Conference on Machine Learning},
	pages={1121--1128},
	year={2009}
}

@article{data_diet,
	title={Deep learning on a data diet: Finding important examples early in training},
	author={Paul, Mansheej and Ganguli, Surya and Dziugaite, Gintare Karolina},
	journal={Advances in Neural Information Processing Systems},
	volume={34},
	pages={20596--20607},
	year={2021}
}

@inproceedings{core-set,
	title={Grad-match: Gradient matching based data subset selection for efficient deep model training},
	author={Killamsetty, Krishnateja and Durga, S and Ramakrishnan, Ganesh and De, Abir and Iyer, Rishabh},
	booktitle={International Conference on Machine Learning},
	pages={5464--5474},
	year={2021},
	organization={PMLR}
}

@inproceedings{ glister,
	title={Glister: Generalization based data subset selection for efficient and robust learning},
	author={Killamsetty, Krishnateja and Sivasubramanian, Durga and Ramakrishnan, Ganesh and Iyer, Rishabh},
	booktitle={Proceedings of the AAAI Conference on Artificial Intelligence},
	volume={35},
	number={9},
	pages={8110--8118},
	year={2021}
}

@inproceedings{clip,
  title={Learning transferable visual models from natural language supervision},
  author={Radford, Alec and Kim, Jong Wook and Hallacy, Chris and Ramesh, Aditya and Goh, Gabriel and Agarwal, Sandhini and Sastry, Girish and Askell, Amanda and Mishkin, Pamela and Clark, Jack and others},
  booktitle={International conference on machine learning},
  pages={8748--8763},
  year={2021},
  organization={PMLR}
}

@article{score-based-3,
  title={What neural networks memorize and why: Discovering the long tail via influence estimation},
  author={Feldman, Vitaly and Zhang, Chiyuan},
  journal={Advances in Neural Information Processing Systems},
  volume={33},
  pages={2881--2891},
  year={2020}
}

@inproceedings{opt-based-1,
  title={PRISM: A Unified Framework of Parameterized Submodular Information Measures for Targeted Data Subset Selection and Summarization},
  author={Kothawade, Suraj and Kaushal, Vishal and Ramakrishnan, Ganesh and Bilmes, Jeff and Iyer, Rishabh},
  booktitle={Thirty-Sixth AAAI Conference on Artificial Intelligence, AAAI},
  year={2022}
}

@inproceedings{opt-based-3,
  title={Coresets for data-efficient training of machine learning models},
  author={Mirzasoleiman, Baharan and Bilmes, Jeff and Leskovec, Jure},
  booktitle={International Conference on Machine Learning},
  pages={6950--6960},
  year={2020},
  organization={PMLR}
}

@inproceedings{opt-based-4,
  title={Submodularity in data subset selection and active learning},
  author={Wei, Kai and Iyer, Rishabh and Bilmes, Jeff},
  booktitle={International conference on machine learning},
  pages={1954--1963},
  year={2015},
  organization={PMLR}
}

@article{forgetting,
	title={An empirical study of example forgetting during deep neural network learning},
	author={Toneva, Mariya and Sordoni, Alessandro and Combes, Remi Tachet des and Trischler, Adam and Bengio, Yoshua and Gordon, Geoffrey J},
	journal={arXiv preprint arXiv:1812.05159},
	year={2018}
}

@inproceedings{
cgscore,
title={Data Valuation Without Training of a Model},
author={Ki Nohyun and Hoyong Choi and Hye Won Chung},
booktitle={The Eleventh International Conference on Learning Representations },
year={2023},
}

@article{gpt-4,
  title={Gpt-4 technical report},
  author={Achiam, Josh and Adler, Steven and Agarwal, Sandhini and Ahmad, Lama and Akkaya, Ilge and Aleman, Florencia Leoni and Almeida, Diogo and Altenschmidt, Janko and Altman, Sam and Anadkat, Shyamal and others},
  journal={arXiv preprint arXiv:2303.08774},
  year={2023}
}

@article{moso,
  title={Data pruning via moving-one-sample-out},
  author={Tan, Haoru and Wu, Sitong and Du, Fei and Chen, Yukang and Wang, Zhibin and Wang, Fan and Qi, Xiaojuan},
  journal={Advances in Neural Information Processing Systems},
  volume={36},
  year={2024}
}

@inproceedings{
beyond,
title={Beyond neural scaling laws: beating power law scaling via data pruning},
author={Ben Sorscher and Robert Geirhos and Shashank Shekhar and Surya Ganguli and Ari S. Morcos},
booktitle={Advances in Neural Information Processing Systems},
editor={Alice H. Oh and Alekh Agarwal and Danielle Belgrave and Kyunghyun Cho},
year={2022},
}

@article{dataset_distillation,
  title={A comprehensive survey to dataset distillation},
  author={Lei, Shiye and Tao, Dacheng},
  journal={arXiv preprint arXiv:2301.05603},
  year={2023}
}

@inproceedings{dataset_distillation2,
  title={Minimizing the accumulated trajectory error to improve dataset distillation},
  author={Du, Jiawei and Jiang, Yidi and Tan, Vincent YF and Zhou, Joey Tianyi and Li, Haizhou},
  booktitle={Proceedings of the IEEE/CVF Conference on Computer Vision and Pattern Recognition},
  pages={3749--3758},
  year={2023}
}

@inproceedings{dataset_distillation3,
  title={Accelerating dataset distillation via model augmentation},
  author={Zhang, Lei and Zhang, Jie and Lei, Bowen and Mukherjee, Subhabrata and Pan, Xiang and Zhao, Bo and Ding, Caiwen and Li, Yao and Xu, Dongkuan},
  booktitle={Proceedings of the IEEE/CVF Conference on Computer Vision and Pattern Recognition},
  pages={11950--11959},
  year={2023}
}

@article{tiny,
	title="A downsampled variant of imagenet as an alternative to the cifar datasets",
	author="Chrabaszcz, Patryk and Loshchilov, Ilya and Hutter, Frank",
	journal="arXiv preprint arXiv:1707.08819",
	month="Aug",
	year="2017"
}

@article{cifar100,
	title={Learning multiple layers of features from tiny images},
	author={Krizhevsky, Alex and Hinton, Geoffrey and others},
	year={2009},
	publisher={Toronto, ON, Canada}
}

@article{yang2024clip,
  title={A CLIP-Powered Framework for Robust and Generalizable Data Selection},
  author={Yang, Suorong and Ye, Peng and Ouyang, Wanli and Zhou, Dongzhan and Shen, Furao},
  journal={arXiv preprint arXiv:2410.11215},
  year={2024}
}

@article{ imagenet-hard,
  title={Imagenet-hard: The hardest images remaining from a study of the power of zoom and spatial biases in image classification},
  author={Taesiri, Mohammad Reza and Nguyen, Giang and Habchi, Sarra and Bezemer, Cor-Paul and Nguyen, Anh},
  journal={Advances in Neural Information Processing Systems},
  volume={36},
  year={2024}
}

@inproceedings{imagenet-a,
  title={Natural adversarial examples},
  author={Hendrycks, Dan and Zhao, Kevin and Basart, Steven and Steinhardt, Jacob and Song, Dawn},
  booktitle={Proceedings of the IEEE/CVF conference on computer vision and pattern recognition},
  pages={15262--15271},
  year={2021}
}

@inproceedings{imagenet-r,
  title={The many faces of robustness: A critical analysis of out-of-distribution generalization},
  author={Hendrycks, Dan and Basart, Steven and Mu, Norman and Kadavath, Saurav and Wang, Frank and Dorundo, Evan and Desai, Rahul and Zhu, Tyler and Parajuli, Samyak and Guo, Mike and others},
  booktitle={Proceedings of the IEEE/CVF international conference on computer vision},
  pages={8340--8349},
  year={2021}
}

@inproceedings{resnet,
	title={Deep residual learning for image recognition},
	author={He, Kaiming and Zhang, Xiangyu and Ren, Shaoqing and Sun, Jian},
	booktitle={Proc. IEEE Conf. Comput. Vis. Pattern Recognit. (CVPR)},
	pages={770--778},
	year={2016}
}

@misc{mscoco,
      title={Microsoft COCO: Common Objects in Context}, 
      author={Tsung-Yi Lin and Michael Maire and Serge Belongie and Lubomir Bourdev and Ross Girshick and James Hays and Pietro Perona and Deva Ramanan and C. Lawrence Zitnick and Piotr Dollár},
      year={2015},
      eprint={1405.0312},
      archivePrefix={arXiv},
      primaryClass={cs.CV},
      url={https://arxiv.org/abs/1405.0312}, 
}

@misc{ade20k,
      title={Semantic Understanding of Scenes through the ADE20K Dataset}, 
      author={Bolei Zhou and Hang Zhao and Xavier Puig and Tete Xiao and Sanja Fidler and Adela Barriuso and Antonio Torralba},
      year={2018},
      eprint={1608.05442},
      archivePrefix={arXiv},
      primaryClass={cs.CV},
      url={https://arxiv.org/abs/1608.05442}, 
}

@article{d2,
  title={D2 pruning: Message passing for balancing diversity and difficulty in data pruning},
  author={Maharana, Adyasha and Yadav, Prateek and Bansal, Mohit},
  journal={arXiv preprint arXiv:2310.07931},
  year={2023}
}

@article{cilimkovic2015neural,
  title={Neural networks and back propagation algorithm},
  author={Cilimkovic, Mirza},
  journal={Institute of Technology Blanchardstown, Blanchardstown Road North Dublin},
  volume={15},
  number={1},
  pages={18},
  year={2015}
}

@misc{ramalingam2023weightedkcenteralgorithmdata,
      title={A Weighted K-Center Algorithm for Data Subset Selection}, 
      author={Srikumar Ramalingam and Pranjal Awasthi and Sanjiv Kumar},
      year={2023},
      eprint={2312.10602},
      archivePrefix={arXiv},
      primaryClass={cs.LG},
      url={https://arxiv.org/abs/2312.10602}, 
}

@article{yang2025rl,
  title={RL-Selector: Reinforcement Learning-Guided Data Selection via Redundancy Assessment},
  author={Yang, Suorong and Li, Peijia and Shen, Furao and Zhao, Jian},
  journal={arXiv preprint arXiv:2506.21037},
  year={2025}
}

@inproceedings{craig,
  title={Coresets for data-efficient training of machine learning models},
  author={Mirzasoleiman, Baharan and Bilmes, Jeff and Leskovec, Jure},
  booktitle={International Conference on Machine Learning},
  pages={6950--6960},
  year={2020},
  organization={PMLR}
}

@inproceedings{crest,
  title={Towards sustainable learning: Coresets for data-efficient deep learning},
  author={Yang, Yu and Kang, Hao and Mirzasoleiman, Baharan},
  booktitle={International Conference on Machine Learning},
  pages={39314--39330},
  year={2023},
  organization={PMLR}
}

@inproceedings{submodular,
  title={Submodular combinatorial information measures with applications in machine learning},
  author={Iyer, Rishabh and Khargoankar, Ninad and Bilmes, Jeff and Asanani, Himanshu},
  booktitle={Algorithmic Learning Theory},
  pages={722--754},
  year={2021},
  organization={PMLR}
}

@misc{liu2023dataefficientaugmentationtrainingneural,
      title={Data-Efficient Augmentation for Training Neural Networks}, 
      author={Tian Yu Liu and Baharan Mirzasoleiman},
      year={2023},
      eprint={2210.08363},
      archivePrefix={arXiv},
      primaryClass={cs.LG},
      url={https://arxiv.org/abs/2210.08363}, 
}

@inproceedings{data-efficient-contrastive-ssl,
  title={Data-efficient contrastive self-supervised learning: Most beneficial examples for supervised learning contribute the least},
  author={Joshi, Siddharth and Mirzasoleiman, Baharan},
  booktitle={International conference on machine learning},
  pages={15356--15370},
  year={2023},
  organization={PMLR}
}

@misc{cazenavette2025datasetdistillationpretrainedselfsupervised,
      title={Dataset Distillation for Pre-Trained Self-Supervised Vision Models}, 
      author={George Cazenavette and Antonio Torralba and Vincent Sitzmann},
      year={2025},
      eprint={2511.16674},
      archivePrefix={arXiv},
      primaryClass={cs.CV},
      url={https://arxiv.org/abs/2511.16674}, 
}

@misc{li2025hyperbolicdatasetdistillation,
      title={Hyperbolic Dataset Distillation}, 
      author={Wenyuan Li and Guang Li and Keisuke Maeda and Takahiro Ogawa and Miki Haseyama},
      year={2025},
      eprint={2505.24623},
      archivePrefix={arXiv},
      primaryClass={cs.LG},
      url={https://arxiv.org/abs/2505.24623}, 
}

@inproceedings{su2024d,
  title={D\^{} 4: Dataset Distillation via Disentangled Diffusion Model},
  author={Su, Duo and Hou, Junjie and Gao, Weizhi and Tian, Yingjie and Tang, Bowen},
  booktitle={Proceedings of the IEEE/CVF Conference on Computer Vision and Pattern Recognition},
  pages={5809--5818},
  year={2024}
}

@inproceedings{dataset_condensation1,
  title={Slimmable dataset condensation},
  author={Liu, Songhua and Ye, Jingwen and Yu, Runpeng and Wang, Xinchao},
  booktitle={Proceedings of the IEEE/CVF Conference on Computer Vision and Pattern Recognition},
  pages={3759--3768},
  year={2023}
}

@article{dataset_condensation2,
  title={An efficient dataset condensation plugin and its application to continual learning},
  author={Yang, Enneng and Shen, Li and Wang, Zhenyi and Liu, Tongliang and Guo, Guibing},
  journal={Advances in Neural Information Processing Systems},
  volume={36},
  year={2023}
}

@article{shao2024elucidating,
  title={Elucidating the design space of dataset condensation},
  author={Shao, Shitong and Zhou, Zikai and Chen, Huanran and Shen, Zhiqiang},
  journal={Advances in Neural Information Processing Systems},
  volume={37},
  pages={99161--99201},
  year={2024}
}

@inproceedings{malakshan2025decomposed,
  title={Decomposed distribution matching in dataset condensation},
  author={Malakshan, Sahar Rahimi and Saadabadi, Mohammad Saeed Ebrahimi and Dabouei, Ali and Nasrabadi, Nasser M},
  booktitle={2025 IEEE/CVF Winter Conference on Applications of Computer Vision (WACV)},
  pages={7112--7122},
  year={2025},
  organization={IEEE}
}

@article{zhong2024dpo,
  title={Dpo meets ppo: Reinforced token optimization for rlhf},
  author={Zhong, Han and Shan, Zikang and Feng, Guhao and Xiong, Wei and Cheng, Xinle and Zhao, Li and He, Di and Bian, Jiang and Wang, Liwei},
  journal={arXiv preprint arXiv:2404.18922},
  year={2024}
}

@misc{schulman2018highdimensionalcontinuouscontrolusing,
      title={High-Dimensional Continuous Control Using Generalized Advantage Estimation}, 
      author={John Schulman and Philipp Moritz and Sergey Levine and Michael Jordan and Pieter Abbeel},
      year={2018},
      eprint={1506.02438},
      archivePrefix={arXiv},
      primaryClass={cs.LG},
      url={https://arxiv.org/abs/1506.02438}, 
}

@misc{yang2025multimodalguideddynamicdatasetpruning,
      title={Multimodal-Guided Dynamic Dataset Pruning for Robust and Efficient Data-Centric Learning}, 
      author={Suorong Yang and Peijia Li and Yujie Liu and Zhiming Xu and Peng Ye and Wanli Ouyang and Furao Shen and Dongzhan Zhou},
      year={2025},
      eprint={2507.12750},
      archivePrefix={arXiv},
      primaryClass={cs.LG},
      url={https://arxiv.org/abs/2507.12750}, 
}

@INPROCEEDINGS{yolov8,
  author={Varghese, Rejin and M., Sambath},
  booktitle={2024 International Conference on Advances in Data Engineering and Intelligent Computing Systems (ADICS)}, 
  title={YOLOv8: A Novel Object Detection Algorithm with Enhanced Performance and Robustness}, 
  year={2024},
  volume={},
  number={},
  pages={1-6},
  doi={10.1109/ADICS58448.2024.10533619}}

@inproceedings{upernet,
  title={Unified perceptual parsing for scene understanding},
  author={Xiao, Tete and Liu, Yingcheng and Zhou, Bolei and Jiang, Yuning and Sun, Jian},
  booktitle={Proceedings of the European conference on computer vision (ECCV)},
  pages={418--434},
  year={2018}
}

@article{ llama,
  title={Llama: Open and efficient foundation language models},
  author={Touvron, Hugo and Lavril, Thibaut and Izacard, Gautier and Martinet, Xavier and Lachaux, Marie-Anne and Lacroix, Timoth{\'e}e and Rozi{\`e}re, Baptiste and Goyal, Naman and Hambro, Eric and Azhar, Faisal and others},
  journal={arXiv preprint arXiv:2302.13971},
  year={2023}
}

@misc{mmlu,
      title={Measuring Massive Multitask Language Understanding}, 
      author={Dan Hendrycks and Collin Burns and Steven Basart and Andy Zou and Mantas Mazeika and Dawn Song and Jacob Steinhardt},
      year={2021},
      eprint={2009.03300},
      archivePrefix={arXiv},
      primaryClass={cs.CY},
      url={https://arxiv.org/abs/2009.03300}, 
}

@misc{alpacaeval,
      title={Length-Controlled AlpacaEval: A Simple Way to Debias Automatic Evaluators}, 
      author={Yann Dubois and Balázs Galambosi and Percy Liang and Tatsunori B. Hashimoto},
      year={2025},
      eprint={2404.04475},
      archivePrefix={arXiv},
      primaryClass={cs.LG},
      url={https://arxiv.org/abs/2404.04475}, 
}

@misc{chen2018gradnormgradientnormalizationadaptive,
      title={GradNorm: Gradient Normalization for Adaptive Loss Balancing in Deep Multitask Networks}, 
      author={Zhao Chen and Vijay Badrinarayanan and Chen-Yu Lee and Andrew Rabinovich},
      year={2018},
      eprint={1711.02257},
      archivePrefix={arXiv},
      primaryClass={cs.CV},
      url={https://arxiv.org/abs/1711.02257}, 
}

@inproceedings{dyn-unc,
  title={Large-scale dataset pruning with dynamic uncertainty},
  author={He, Muyang and Yang, Shuo and Huang, Tiejun and Zhao, Bo},
  booktitle={Proceedings of the IEEE/CVF Conference on Computer Vision and Pattern Recognition},
  pages={7713--7722},
  year={2024}
}

@inproceedings{
greats,
title={{GREATS}: Online Selection of High-Quality Data for {LLM} Training in Every Iteration},
author={Jiachen T. Wang and Tong Wu and Dawn Song and Prateek Mittal and Ruoxi Jia},
booktitle={The Thirty-eighth Annual Conference on Neural Information Processing Systems},
year={2024},
url={https://openreview.net/forum?id=232VcN8tSx}
}

@misc{opus,
      title={OPUS: Towards Efficient and Principled Data Selection in Large Language Model Pre-training in Every Iteration}, 
      author={Shaobo Wang and Xuan Ouyang and Tianyi Xu and Yuzheng Hu and Jialin Liu and Guo Chen and Tianyu Zhang and Junhao Zheng and Kexin Yang and Xingzhang Ren and Dayiheng Liu and Linfeng Zhang},
      year={2026},
      eprint={2602.05400},
      archivePrefix={arXiv},
      primaryClass={cs.CL},
      url={https://arxiv.org/abs/2602.05400}, 
}

@misc{wang2026winningpruninggambleunified,
      title={Winning the Pruning Gamble: A Unified Approach to Joint Sample and Token Pruning for Efficient Supervised Fine-Tuning}, 
      author={Shaobo Wang and Jiaming Wang and Jiajun Zhang and Cong Wang and Yue Min and Zichen Wen and Xingzhang Ren and Fei Huang and Huiqiang Jiang and Junyang Lin and Dayiheng Liu and Linfeng Zhang},
      year={2026},
      eprint={2509.23873},
      archivePrefix={arXiv},
      primaryClass={cs.CL},
      url={https://arxiv.org/abs/2509.23873}, 
}

@article{dual,
  title={Lightweight dataset pruning without full training via example difficulty and prediction uncertainty},
  author={Cho, Yeseul and Shin, Baekrok and Kang, Changmin and Yun, Chulhee},
  journal={arXiv preprint arXiv:2502.06905},
  year={2025}
}
\bibliographystyle{icml2026}


\end{document}